\documentclass[runningheads]{llncs}
\usepackage{graphicx}
\usepackage{soul, color}
\begin{document}
%

\title{Detecting Mitoses with a Convolutional Neural Network for MIDOG 2022 Challenge}
%
%
\author{Hongyan Gu\inst{1}\orcidID{0000-0001-8962-9152} \and
Mohammad Haeri\inst{2}\orcidID{0000-0001-6055-9779} \and
Shuo Ni\inst{1}\orcidID{0000-0002-4125-2373} \and 
Christopher Kazu Williams\inst{3}\orcidID{0000-0003-3253-7630} \and
Neda Zarrin-Khameh\inst{4}\orcidID{0000-0001-7506-156X} \and
Shino Magaki\inst{3}\orcidID{0000-0003-0433-5759} \and
Xiang `Anthony' Chen\inst{1}\orcidID{0000-0002-8527-1744}
}
\authorrunning{Gu et al.}
%
\institute{University of California, Los Angeles, USA \and
University of Kansas Medical Center, USA \and 
UCLA David Geffen School of Medicine, USA \and
Baylor College of Medicine, USA \\
\email{xac@ucla.edu}}
\maketitle              
\begin{abstract}
This work presents a mitosis detection method with only one vanilla Convolutional Neural Network (CNN). Our method consists of two steps: given an image, we first apply a CNN using a sliding window technique to extract patches that have mitoses; we then calculate each extracted patch's class activation map to obtain the mitosis's precise location. To increase the model performance on high-domain-variance pathology images, we train the CNN with a data augmentation pipeline, a noise-tolerant loss that copes with unlabeled images, and a multi-rounded active learning strategy. In the MIDOG 2022 challenge, our approach, with an EfficientNet-b3 CNN model, achieved an overall F1 score of 0.7323 in the preliminary test phase, and 0.6847 in the final test phase (task 1). Our approach sheds light on the broader applicability of class activation maps for object detections in pathology images.


\end{abstract}

\section{Introduction}

Mitotic activity is a crucial histopathological indicator related to cancer malignancy and patients' prognosis \cite{cree2021counting}.  Because of its importance, a considerable amount of literature has proposed datasets \cite{aubreville2020completely,bertram2019large} and deep learning models \cite{li2018deepmitosis,mahmood2020artificial} for mitosis detection. To date, a number of state-of-the-art deep learning approaches detect mitoses with the segmentation task \cite{jahanifar2021stain,yang2021sk}. However, these 
methods usually require generating pixel-level segmentation maps as ground truth. Another school of techniques utilizes object detection with a two-stage setup --- a localization model (\textit{e.g.,} RetinaNet) is employed for extracting interest locations, followed by a classification model to tell whether these locations have mitoses \cite{aubreville2020completely,li2018deepmitosis,mahmood2020artificial}. Such a two-stage setup was reported to improve the performance of mitosis detection compared to that with the localization model only \cite{aubreville2020completely}.

Since adding a classification model can improve the performance, we argue that using only one CNN model for mitosis detection is also viable. Because CNNs cannot directly report the location of mitosis, previous works either modified the structure of CNNs \cite{cirecsan2013mitosis}, or used CNNs with a small input size to reduce the localization errors \cite{8327641}. Instead, our approach extracts the location of mitoses with the class activation map (CAM) \cite{zhou2016learning}, which allows CNNs to accept a larger input size for more efficient training. Also, our approach is model-agnostic, and can work with vanilla CNNs because calculating CAMs does not require changing the network structure. 

We validated our proposed method in MItosis DOmain Generalization (MIDOG) 2022 Challenge \cite{midog2022}. The challenge training set consists of 403 Hematoxylin \& Eosin (H\&E) regions of interest (ROIs, average size=$5143\times 6860$ pixels), covering six tumor types scanned from multiple scanners. 354/403 ROIs have been labeled and have 9,501 mitotic figures. The preliminary test set includes 20 cases from four tumor types, and the final test set has 100 independent tumor cases from ten tumor types. Given the dataset's high variance, we employed three techniques to improve the CNN's performance:

\begin{enumerate}
    \item An augmentation pipeline with balanced-mixup \cite{galdran2021balanced} and stain augmentation \cite{8327641};
    \item The utilization of unlabeled images for training (treated as negative), and training with the Online Uncertainty Sample Mining (OUSM) \cite{xue2019robust} to gain robustness with noisy labels;
    \item A multi-rounded, active learning training strategy that adds false-positive, false-negative, and hard-negative patches after each round of training.
\end{enumerate}

\section{Methods}

\subsection{Extracting Patches for Initial Training}

We randomly used $\sim 90\%$ of the image instances in the MIDOG 2022 Challenge to generate the training set and $\sim 10\%$ for the validation set. To maximally utilize the dataset, we included unlabelled images in the training set and treated them as negative images (\textit{i.e.,} no mitoses inside). For each image, we extracted patches with the size of $240\times 240 \times 3$ pixels surrounding the location of each annotation (provided by the challenge) and placed them into the train/validation set.

\begin{figure}
\centering
\includegraphics[width=\linewidth]{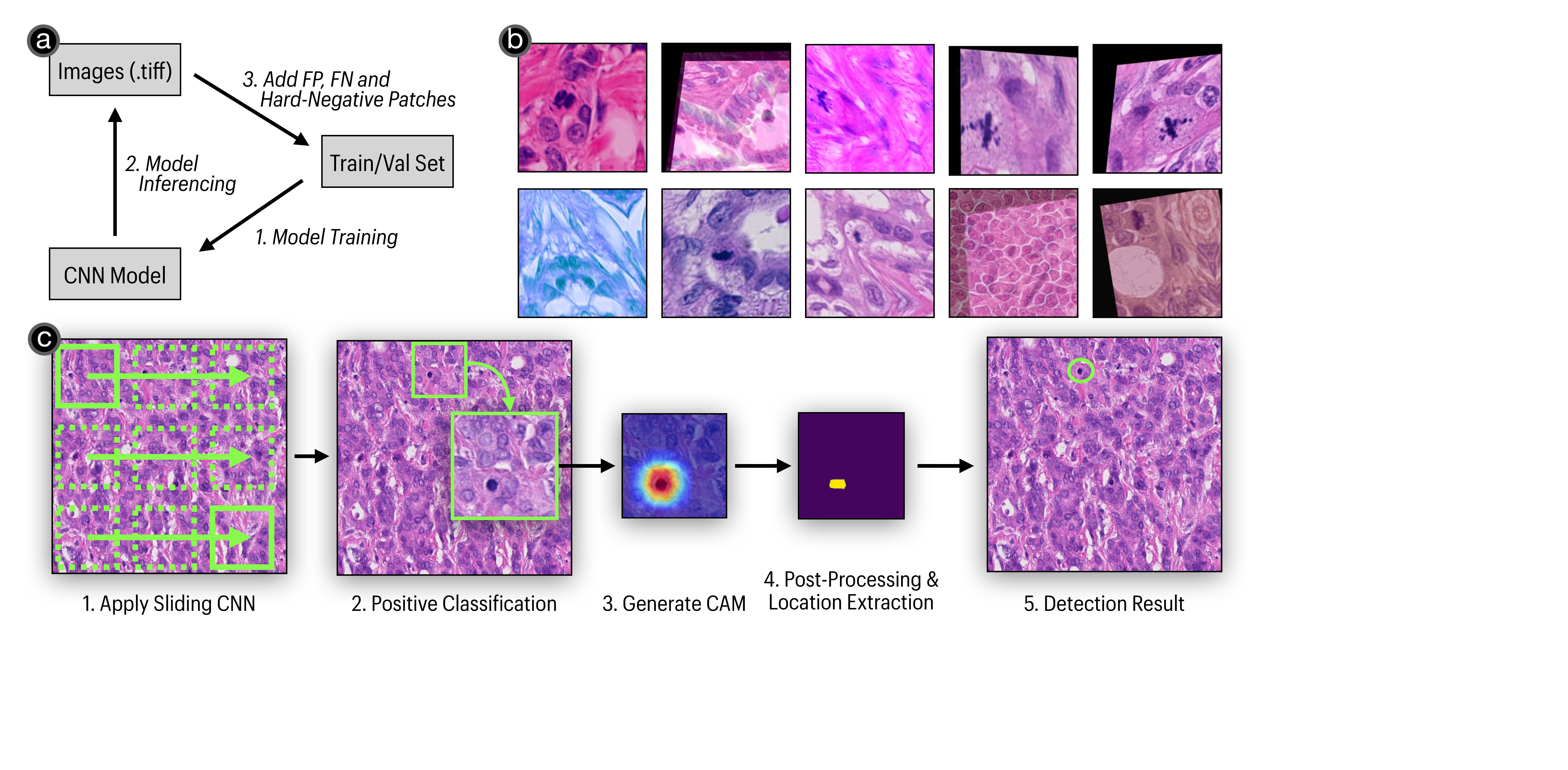}
\caption{(a) Illustration of the active learning training strategy used in this work; (b) Examples of augmented patches according to our augmentation pipeline; (c) Overall data processing pipeline our approach: detecting mitosis using a convolution neural network and the class activation map.}
\label{fig1}
\end{figure}

\subsection{Model Training}
\label{sec:train}
We trained an EfficientNet-b3 \cite{tan2019efficientnet} model (input size: $240\times 240 \times 3$) with pre-trained ImageNet weights. Here, we tried to improve model performance by constructing an online data augmentation pipeline. The pipeline includes general image augmentation techniques, including random rotation, flip, elastic transform, grid distortion, affine, color jitter, Gaussian blur, and Gaussian noise. Besides, we added two augmentation methods -- stain augmentation \cite{8327641} and balance-mixup \cite{galdran2021balanced} -- to deal with the domain shift in pathology images. Examples of augmented patches are shown in Figure \ref{fig1}(b). The model was trained with an SGD optimizer with momentum 0.9, a Cosine Annealing learning rate scheduler with warm restart (max LR=$6\times 10^{-4}$). Since we treated all unlabeled images as negative, we further used an OUSM\cite{xue2019robust} + COnsistent RAnk Logits (CORAL) loss \cite{cao2020rank} to deal with noisy labels. Each round of training had 100 epochs, and we selected the model with the highest F1 score on the validation set for inferencing.

\subsection{Inferencing}
\label{sec:inf}
We slid the trained EfficientNet on train and validation images with window size $240\times 240$ and step-size 30. We then cross-referenced the CNN predictions with the ground truth. Here, we define a positive window classification as a true-positive if mitoses were inside the window and a false-positive otherwise. We further define false-negative if no positive windows surround a mitosis annotation.

\subsection{Incrementing the Patch Dataset with Active Learning}
\label{sec:act}
We employed a multi-round active learning process to boost the performance of the EfficientNet (Figure \ref{fig1}(a)). Each round starts with the model training on the current train/validation set (Section \ref{sec:train}). Then, the best model is selected and applied to the images (Section \ref{sec:inf}). After that, false-positive, false-negative, and hard-negative patches are added to the train/validation set. The procedure was repeated six times until the model's F1 score on the validation set does not increase. Eventually, there are 103,816 patches in the final training set, and 23,638 in the validation set.

\subsection{Extracting Mitosis Locations with CAMs} 
\begin{figure}
\centering
\includegraphics[width=\linewidth]{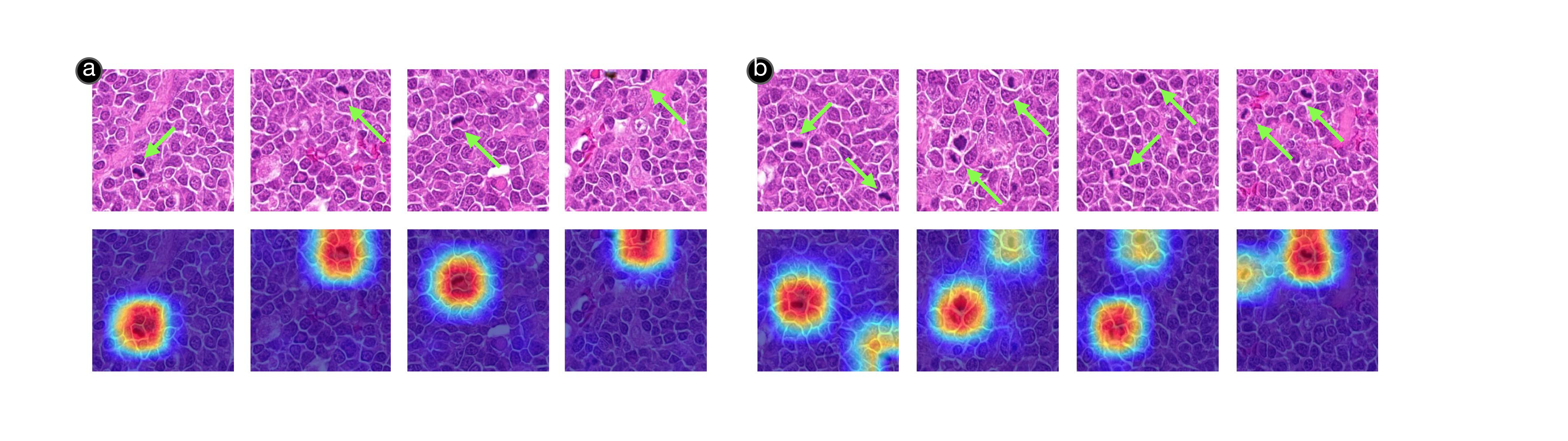}
\caption{Examples of positive patches detected by CNN and corresponding CAMs generated by GradCAM++ from image \texttt{240.tiff}. (a) Patch samples and CAMs if there is one mitosis inside (pointed by the green arrow); (b) Patch samples and CAMs if there are multiple mitoses inside. Note that generated CAMs might not report strong signals for some mitoses.}
\label{fig2}
\end{figure}

We used the best model from the final round in Section \ref{sec:act} for the test images. A window with a CNN probability $>0.84$ was considered positive, and non-maximum suppression with a threshold of 0.22 was used to mitigate the overlapping windows. For each positive window, we calculated the CAM with GradCAM++ \cite{gradcam++}, and extracted the hotspot's centroid as the mitosis location (Figure \ref{fig1}(c)). Examples of CAMs are shown in Figure \ref{fig2}(a). Specific numbers and thresholds were selected according to the best F1 performance from the validation images.

\section{Results}
On the preliminary test phase of the MIDOG 2022 Challenge, our approach achieved an overall F1 score of 0.7323, with 0.7313 precision and 0.7333 recall. In the final test phase of the challenge (task 1), our approach achieved the F1 score of 0.6847 (precision:0.7559, recall: 0.6258). In sum, our approach is $2.34\%$ higher than the baseline RetinaNet approach regarding the overall F1 score in the preliminary test phase, but $4.21\%$ lower in the final test phase. Please refer to the Grand-Challenge Leader-board\footnote{\url{https://midog2022.grand-challenge.org/evaluation/final-test-phase-task-1-without-additional-data/leaderboard/}} for more details of the test result.

\section{Discussion \& Conclusion}
Although CAMs are primarily used for explaining CNN classifications, we demonstrate their potential usage to help detect mitoses in H\&E images. We believe our approach has the potential to work with mobile/edge devices, which have limited computational power and are not optimized for special-structured deep-learning models.

It is noteworthy that CAMs might fail to highlight all mitoses when there are multiple in an image. As shown in Figure \ref{fig2}(b), if there were multiple mitoses inside one positive patch, CAMs might report a strong signal for one mitosis, while giving weak signals for others. However, we argue that our approach \textit{as-is} can still be helpful in diseases where mitosis is not a high-prevalent histological pattern (\textit{e.g.,} meningiomas \cite{gu2021xpath}).

On the other hand, we believe that the limitation is partly caused by using classification models for localization tasks: given a patch with multiple mitoses, a CNN only needs to find at least one to predict it as positive. As such, other mitoses are more or less ignored. To compensate for the limitation, we suggest future works to improve the quality of CAMs by aligning them with ground-truth mitosis location maps. For example, a ``CAM-loss'' might be designed to penalize the misalignment between the CAM and mitosis location heatmaps \cite{zhu2022gaze}. Different from the segmentation task, generating mitosis location maps does not require pixel-level segmentation masks: simply applying a Gaussian kernel over the locations of mitoses can generate a mitosis location map similar to Figure \ref{fig2}.

\section{Acknowledgement}
This work was funded in part by the Office of Naval Research under grant N000142212188.

%
%
%
\bibliographystyle{splncs04}
\bibliography{literature}
\end{document}